\documentclass[journal]{IEEEtran}

\usepackage{amsmath}
\usepackage{amssymb}

\allowdisplaybreaks 

\usepackage{array}
\usepackage{multirow}
\usepackage{multicol}
\usepackage{graphicx}
\usepackage{adjustbox}
\usepackage{algorithm}
\usepackage{algorithmic}
\usepackage{pifont}
\usepackage{xcolor}
\usepackage{url}
\usepackage[hidelinks]{hyperref}

\hypersetup{%
	pdfauthor={Yusen Wan, Zeyuan Chen, Qianshi Zou, Xu Chen},
	pdftitle={R-SLPR: Region-based Small-to-Large Point-cloud Registration with Contrastive Learning},
	pdfkeywords={Point-cloud Registration, Small-to-Large Alignment, Contrastive Learning},
	pdfsubject={Region-based Small-to-Large Point-cloud Registration with Contrastive Learning},
}

\begin{document}

\title{R-SLPR: Region-based Small-to-Large Point-cloud Registration with Contrastive Learning}

\author{Yusen~Wan, Zeyuan~Chen, Qianshi~Zou, and Xu~Chen
\thanks{Yusen Wan, Zeyuan Chen, Qianshi Zou, and Xu Chen are with the Department
of Mechanical Engineering, University of Washington, Seattle, WA, USA.}}

\maketitle


\begin{abstract}
Point-cloud (PC) registration is fundamental to three-dimensional (3D) perception in robotic systems. 
However, classic registration algorithms falter when aligning a source PC containing limited, incomplete, or ambiguous geometric cues against a reference. 
This challenge of registering a small, partial PC to a significantly larger global reference is pervasive in real-world deployment yet remains insufficiently addressed by existing learning-based approaches, which typically assume comparable scales and significant overlap. 
To bridge this gap, we propose the Region-based Small-to-Large Point-cloud Registration framework (R-SLPR), a novel three-stage architecture that fundamentally reformulates the scale-mismatched registration problem into a sequence of region proposal, regional matching, and iterative refinement. 
Unlike conventional methods that fail to localize specific regions, R-SLPR explicitly identifies candidate regions prior to estimating rigid transformations, ensuring robust alignment even under severe scale mismatch.
The framework introduces a Fibonacci Grid Segmentation method coupled with a contrastive learning objective to effectively generate and match local geometric patches. Building on this, a novel Cascade Anchor Selection and Refinement algorithm iteratively aligns the source with the target region to maximize precision.
Extensive evaluation on ModelNet40 demonstrates that R-SLPR establishes a new state-of-the-art accuracy standard, outperforming prior approaches and significantly reducing position and rotation Mean Absolute Error (MAE) to 0.009 and 1.104, respectively. For more details and supplementary materials, please visit the companion website of this paper\footnote{\url{https://macs-lab.github.io/R-SLPR/}}.
\end{abstract}


\section{Introduction}
\label{sec:introduction}
The rise of machine vision and 3D imaging has created significant opportunities for 3D machine perception with point clouds (PCs) in automation and robotics \cite{cite1, cite2, cite3}. Robots can capture PCs of surrounding environments constantly and leverage registration methods to align and fuse these PCs to build a complete scene PC---a collection of 3D data that provides spatial information for such downstream tasks as navigation and manipulation. In this process, PC registration is crucial for aligning different spatial data, or more specifically, for providing the rigid transform between PCs and determining the fusion result of the final scene PC.

\begin{figure}[!t]
\centerline{\includegraphics[width=\columnwidth]{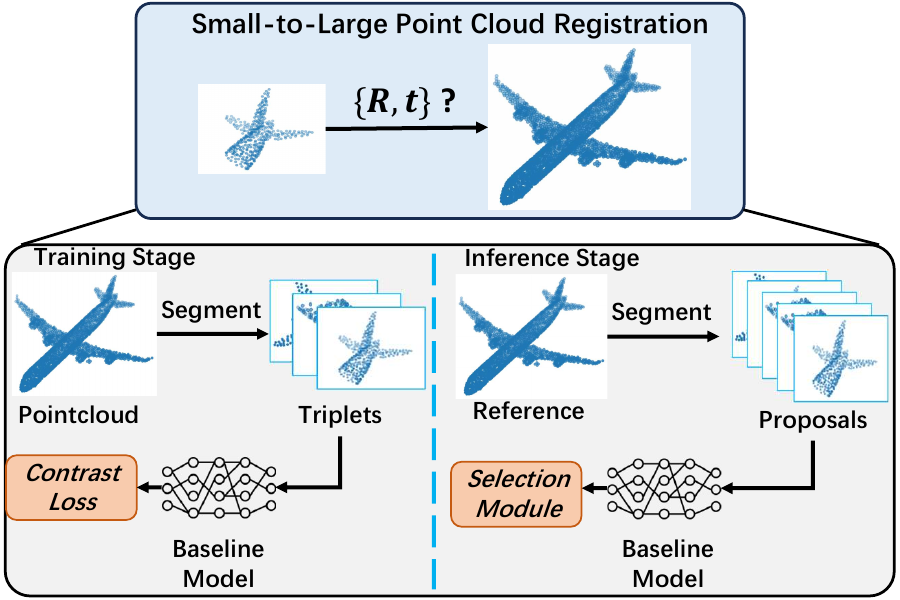}}
\caption{Example illustration of the small-to-large PC registration problem and an overview of the proposed method.}
\label{graphic_abstract}
\end{figure}

Classic PC registration has leveraged iterative local optimization and hand-crafted geometric descriptors, such as the Iterative Closest Point (ICP) algorithm \cite{besl1992method} and Fast Point Feature Histograms (FPFH) \cite{rusu2009fast}. 
Recent advances in PC registration have capitalized on deep neural networks (DNNs): e.g., RPMNet \cite{RPMNet} for robust correspondence learning via Sinkhorn layers, PointNetLK \cite{PointNetLK} for iterative Lucas-Kanade alignment, and MCLNet \cite{MCLNet} for multi-scale feature consistency.
These approaches are typically trained and evaluated using a source–reference split on multiple PC datasets, such as ModelNet40 \cite{ModelNet40} or KITTI \cite{geiger2013ijrr}.
More specifically, the reference PCs first undergo randomly generated rigid transformations that serve as ground-truth (GT) transforms. 
The networks then extract features and estimate the rigid transformation between the source and reference PCs, and compare the estimated transforms with the GT transforms for model training and performance evaluation. 
For this workflow to work, the source and reference PCs must maintain \emph{comparable scales, substantial overlap, and congruent geometric coverage}.

However, as shown in Fig. \ref{graphic_abstract}, such assumptions are routinely violated in such real-world scenarios as manufacturing, where the source and reference PCs often differ drastically in size, density, and completeness, resulting in severe geometric and statistical mismatches \cite{S-L 1, S-L 2}. For example, in computer-assisted orthopedic surgery (CAOS), partially acquired PCs may capture as little as 20\% of the target anatomy \cite{S-L 1}, while in computer-assisted manufacturing, workpiece localization typically requires aligning a sparse, partial scan with a complete PC derived from a CAD model \cite{iLSPR}. In these settings, the overlap between the two PCs is not only limited but also often unknown \emph{a priori}, leading to what we shall refer to as the \emph{small-to-large PC registration problem}. 

At the core of small-to-large PC registration is the challenge of identifying the correct region of a much larger reference point cloud when the source contains only a small and incomplete fragment. To address this fundamental issue, we propose a first-of-its-kind Region-based Small-to-Large Point Cloud Registration framework (R-SLPR), which reformulates registration as a region propose–match–refine process---a process that equips the system with an explicit region identification ability before estimating the rigid transformations, thereby enabling accurate registration even under severe scale mismatch. More specifically, to provide full and unbiased coverage of the reference point cloud, we first introduce a uniform segmentation strategy based on Fibonacci sphere sampling, which generates deterministic and evenly distributed directional anchors for constructing region proposals. To further strengthen the model’s ability to distinguish relevant regions from irrelevant ones, we design a patch-level contrastive learning strategy that pulls together embeddings of geometrically corresponding regions while pushing apart those of non-overlapping regions. In addition, a lightweight Cascade Anchor Selection and Refinement algorithm (CASR) iteratively evaluates qualities of the region proposals using alignment distance and gradually updates anchor directions, allowing the system to converge toward the true corresponding region even when the initial proposal is coarse. Extensive experiments on ModelNet40 demonstrate that R-SLPR substantially enhances registration robustness and accuracy under clean, noisy, unseen-category, and mixed conditions, achieving a position Root Mean Square Error (RMSE) of 0.02 and a rotation RMSE of 2.6 in benchmark evaluations.

Our main contributions are summarized as follows:
\begin{itemize}
    \item We present R-SLPR, a first-of-its-kind region-based registration framework that enables reliable small-to-large alignment by structuring the process into proposal generation, matching, and refinement.

    \item We propose a deterministic Fibonacci Grid Segmentation strategy that yields directionally uniform region proposals over the reference point cloud.

    \item We introduce a patch-level contrastive learning objective that enhances the discriminability of regional feature embeddings under severe scale disparity.

    \item We develop a Cascade Anchor Selection and Refinement algorithm that iteratively improves proposal quality and converges toward the correct region with minimal computational overhead.

    \item We validate the effectiveness of R-SLPR through extensive experimentation, demonstrating consistent improvements over representative registration baselines across diverse conditions.

\end{itemize}

The remainder of this article is organized as follows. Section \ref{related works} describes the background and related work. Section \ref{Problem Formulation} introduces the problem formulation. Section \ref{proposed method} introduces the proposed R-SLPR method. Section \ref{sec:experimental evaluation} introduces the evaluation experiments. Section \ref{sec: conclusion and future work} concludes the paper.

\section{Related Works}
\label{related works}
PC registration aims to estimate the rigid transformation between two PCs. A classical solution is the Iterative Closest Point (ICP) algorithm, which iteratively establishes point correspondences via nearest-neighbor search and estimates the rigid transformation using singular value decomposition (SVD) \cite{besl1992method}. While effective in controlled settings, ICP is sensitive to initialization, noise, outliers, and limited overlap.

With the advancement of deep learning, numerous learning-based PC registration methods have been proposed. These methods can be broadly categorized into correspondence-based approaches and global feature-based approaches. Correspondence-based methods currently dominate the literature. To overcome the limitations of ICP, Wang \emph{et al.} proposed Deep Closest Point (DCP), which replaces hand-crafted descriptors with deep feature extraction and learns point correspondences using a neural network \cite{DCP}. To further improve correspondence reliability under noise and outliers, Wang \emph{et al.} introduced the Multi-Features Guide Network (MFGNet), which employs a keypoint selection module to filter outliers and computes matching matrices using multiple learned features \cite{MFGNet}. Similarly, Ginzburg \emph{et al.} addressed the influence of outliers by defining a cosine-likelihood--based soft sampling map and selecting the top-$K$ most confident points for rigid transformation estimation \cite{DWC}. In this framework, features are extracted using modified Dynamic Graph CNNs (DGCNNglob and DGCNNloc) derived from DGCNN \cite{DGCNN}. In scenarios where explicit one-to-one correspondences are ambiguous, alternative strategies have been explored. For instance, RPMNet replaces hard correspondence selection with a weighted average of potential matches by predicting a soft matching matrix and transformation parameters using a deep network \cite{RPMNet}. In addition to correspondence-based methods, several approaches rely on global feature representations. Sarode \emph{et al.} proposed PointNetLK, which uses PointNet to extract global features and applies the Lucas--Kanade algorithm to estimate the rigid transformation \cite{PointNetLK}. Li \emph{et al.} further improved its generalization ability in PointNetLK-Revisited \cite{PointNetLK-revisited}. Wu \emph{et al.} proposed RORNet, a dual-branch overlap estimation network that filters points with low inlier probability, effectively combining similarity-based and score-based registration strategies \cite{RORNet}. Similarly, Tan \emph{et al.} introduced MCLNet, an end-to-end framework that enforces point-level consistency for filtering and correspondence-level consistency for reliable matching matrix estimation \cite{MCLNet}.

Although these methods achieve high registration accuracy and computational efficiency, they primarily focus on registering PCs of comparable size with moderate to high overlap. The problem of aligning PCs with significant size disparities remains largely unaddressed. Chen \emph{et al.} explored a full-to-partial registration framework based on reinforcement learning and evaluated robustness across different overlap ratios \cite{S-L 1}; however, their work focuses on improving robustness rather than explicitly addressing the fundamental challenges posed by extreme size asymmetry. Overall, despite its prevalence in real-world applications, the \emph{small-to-large PC registration problem} remains insufficiently explored in the existing literature.

\section{Problem Formulation}
\label{Problem Formulation}
\subsection{Small-to-Large Point Cloud Registration}
\label{sec:slpcr}
Consider two PCs with a significant disparity in scale. We denote the smaller fragment as the source PC $S = \{s_i\in \mathbb{R}^3 \mid i=1,2,\ldots,I\}$ and the complete PC as the reference PC $Q=\{q_j\in \mathbb{R}^3 \mid j=1,2,\ldots,J\} ,$ where $I\ll J .$ To quantify the scale mismatch, we define $\gamma \triangleq \frac{I}{J}$ as the cardinality ratio between PCs to be aligned.
We aim to estimate the rigid transformation $\{R,t\}$ between these two PCs, where $R \in SO(3)$ is a rotation matrix and $t \in \mathbb{R}^3$ is a translation vector. 

To show the correspondence of points, we define a matching matrix $M=\{m_{ij} \in[0,1] \mid i=1,2,\ldots,I;\ j=1,2,\ldots,J\} .$ The objective of small-to-large point cloud registration is to determine the optimal transformation that minimizes the alignment error between the transformed source fragment and the reference model. This is mathematically formulated as minimizing the mean distance between the transformed source PC $S$ and the reference PC $Q :$
\begin{equation}
\label{eq1}
\arg\min_{R,t,M} \sum_{i=1}^I \sum_{j=1}^J m_{ij} \parallel Rs_i+t-q_j \parallel_2  
\end{equation}
The value of $m_{ij}$ here represents the degree of matching between $s_i$ and $q_j$, and satisfies: $\forall i ,$ $\sum_{j=1}^J m_{ij} =1 .$ This summation holds because every point $s_i \in S$ corresponds to exactly one point in $Q .$

\subsection{Baseline Iterative Corresponding Point Match Method}

Before we introduce the proposed solution approach, we summarize the core iterative refinement solution structure involved. 

Mainstream correspondence-based methods rely on iterative point cloud matching. Each iteration consists of three primary stages: feature extraction, correspondence estimation, and transformation recovery.
For each iteration, the source PC $S$ and the reference PC $Q$ form as the inputs, and the rigid transformation $\{R,t\}$ is the output. 
First, $S$ and $Q$ are fed into a feature extraction module to obtain the deep features $F_S$ and $F_Q$ by $F_S=f(S)$ and $F_Q=f(Q) .$ 
Then, the deep features and the original coordinates will be used for the matching-matrix calculation, mathematically abstracted as $M=g(F_S,F_Q,S,Q)$.

After obtaining the matching matrix, the corresponding coordinate $\hat{q}_i$ in the reference PC can be aligned to each point $s_i$ in the source PC by two methods. The first method is a weighted average:
\begin{equation}
\label{eq2}
\hat{q}_i = \frac{1}{\sum_{j=1}^n m_{ij}} \sum_{j=1}^n (m_{ij} \cdot q_j)
\end{equation}
The second method is the selection of the highest score by $\hat{q}_i = q_j ,$ where $j = \max_j m_{ij} .$ 

After getting the corresponding coordinate, we can calculate the rigid transforms between $S$ and $Q$ by SVD. 
Specifically, we build two coordinate matrices $ \hat{S} = [s_1^T, s_2^T, \ldots, s_I^T] \in R^{3 \times I}$ and $ \hat{Q} = [\hat{q}_1^T, \hat{q}_2^T,\ldots, \hat{q}_I^T] \in R^{3 \times I}$ to compute $H=\hat{S} \hat{Q}^T .$ 
Computing the SVD of $H$, $H=U \Sigma V^T$, yields the rotation matrix $R=VU^T$ and the translation vector $t = \frac{1}{n} \sum_{i=1}^I \{\hat{q}_i - Rs_i \} .$ 

\section{Region-based Small-to-Large PC Registration Method} 
\label{proposed method}
\subsection{Overview}

\begin{figure*}[!t]
\centerline{\includegraphics[width=\linewidth]{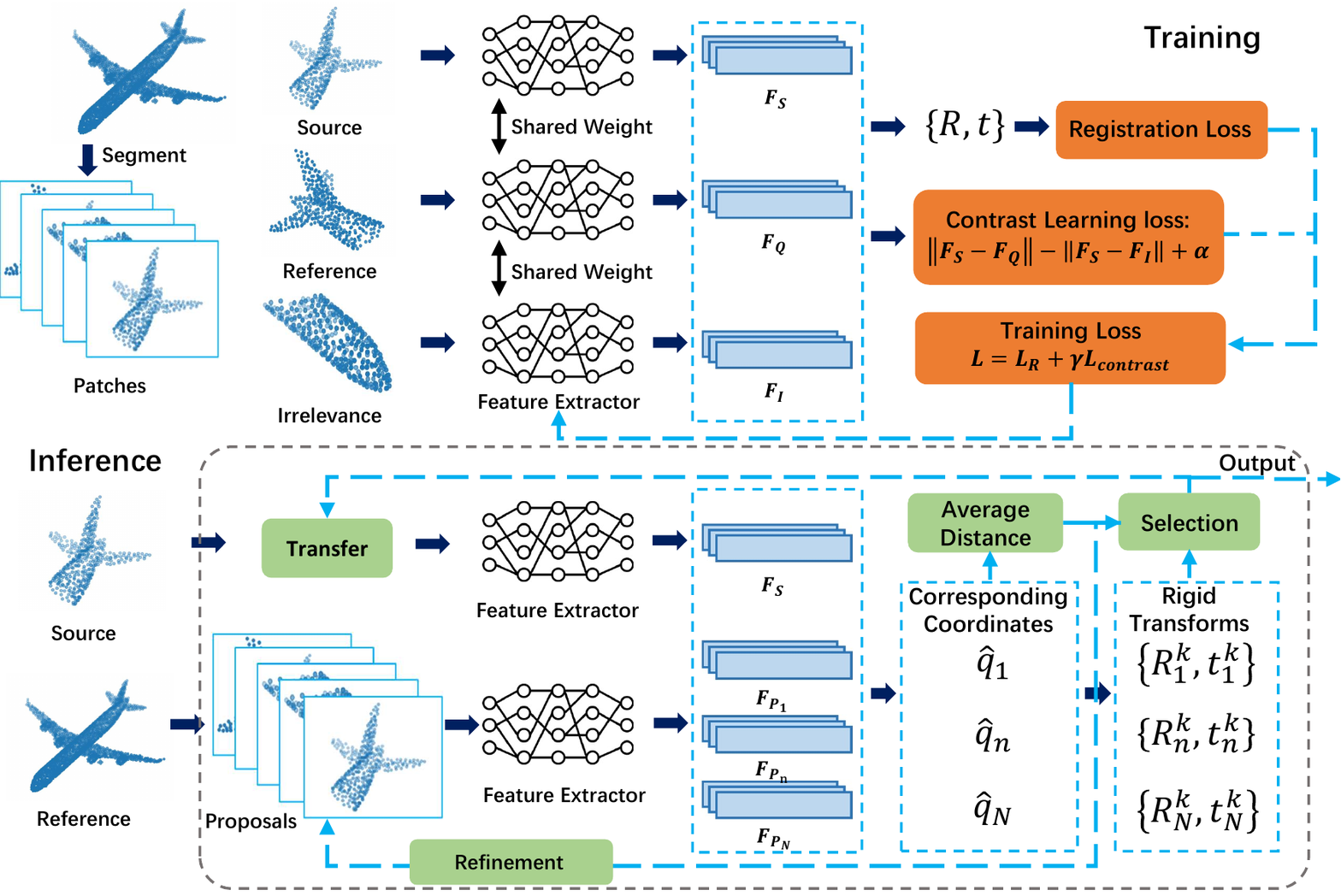}}
\caption{The overall structure of the proposed R-SLPR method.}
\label{overall_process}
\end{figure*}

Fig. \ref{overall_process} shows the proposed R-SLPR framework to address the small-to-large PC matching problem. 
At the core, R-SLPR introduces a region identification before estimating the rigid transformations, and forms a contrastive learning in a two-stage training-inference sequence. 
This is achieved algorithmically by integrating a baseline registration network, a contrastive learning loss function, a Fibonacci Grid Segmentation (FGS) strategy for region proposal generation, and a Cascade Anchor Selection and Refinement (CASR) module for iterative proposal refinement. 

Given a small source PC fragment and a much larger reference PC, R-SLPR first decomposes the reference PC into a set of overlapping regional proposals $P = \{P_n \mid n = 1, 2,\ldots, N\}$ using the FGS scheme, where each proposal is associated with a deterministic anchor direction.

In the training stage, we introduce three PCs: a source PC $S$, a reference PC $Q$ from an overlapping proposal, and an irrelevance PC $Ir$ from a distant region. The goal of the training stage is to regress accurate rigid transformations between overlapping patches and produce discriminative feature embeddings that separate relevant and irrelevant regions by minimizing the $L_2$ distance $\parallel f(S)-f(Q) \parallel$ and maximizing $\parallel f(S)-f(Ir) \parallel .$


In the inference stage, the trained baseline model is applied to every proposal in the current proposal set. For each proposal, the model predicts a local rigid transformation between $S$ and the proposal $P = \{P_n \mid n = 1, 2,\ldots, N\} .$ The prediction results are then evaluated by an average alignment distance metric, according to which R-SLPR selects the best anchor by a CASR module and updates all anchors via an exponential moving average towards the best anchor. By repeating segmentation and selection with the updated anchors, R-SLPR gradually converges to the correct region and the final rigid transformation.


\subsection{Model Training}
\subsubsection{Fibonacci Grid Segmentation and Triplet Generation}
\begin{figure}[!t]
    \centering
    \includegraphics[width=\columnwidth]{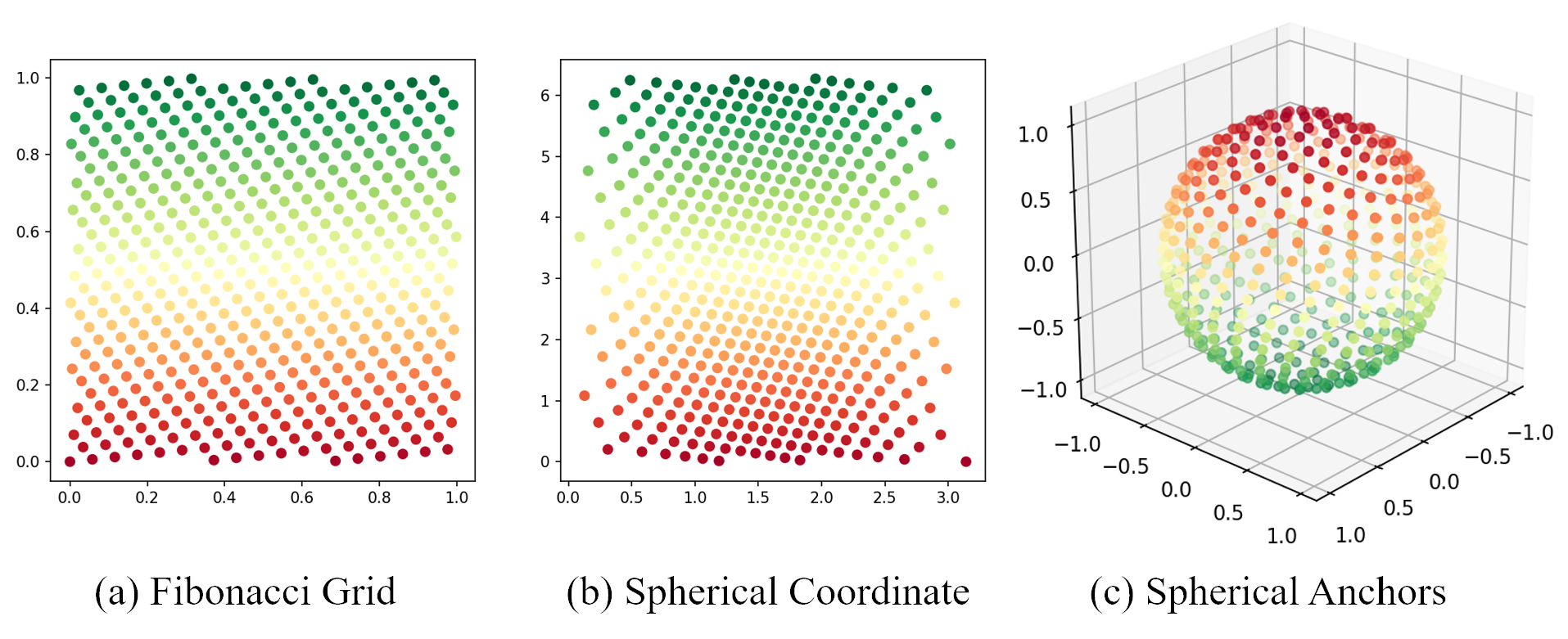}
\caption{Proposed anchor generation based on a Fibonacci grid: (a) the original grid, (b) the spherical coordinates, and (c) the anchors.}
\label{fig_fibonacci}
\end{figure}

\begin{figure}[!t]
\centerline{\includegraphics[width=\columnwidth]{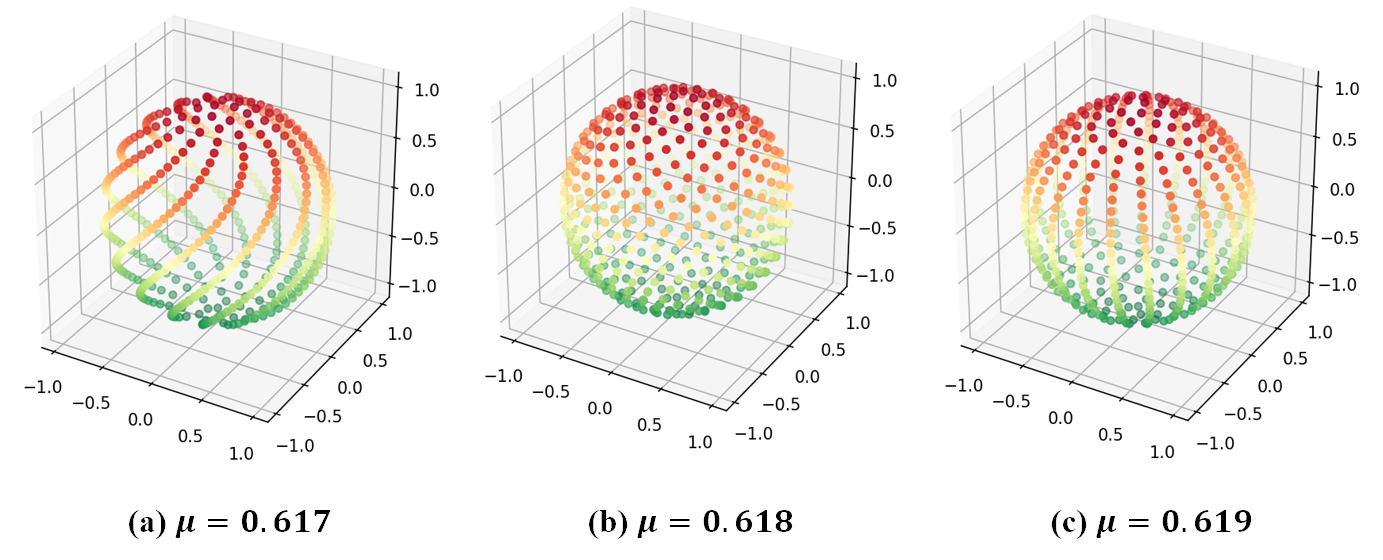}}
\caption{Effect of $\mu$ in the proposed anchor generation.}
\label{miu_value}
\end{figure}

Because input PCs are sampled on surfaces rather than volumes, voxel-based methods cannot guarantee uniform and complete coverage of PCs. 
Instead, we propose a Fibonacci Grid Segmentation (FGS) method and leverage a Fibonacci grid, a deterministic lattice, to generate directionally uniform anchors on the unit sphere \cite{fibonacci} and use them to segment the PCs into patches fully and uniformly (Fig. \ref{fig_fibonacci}).

First, we generate a Fibonacci grid as illustrated in Fig. \ref{fig_fibonacci}a, based on the following coordinate definition:
\begin{equation}
\label{eq3}
    \begin{cases}
        u_n=\frac{n-1}{N} \\
        v_n = \{ \mu n\}
    \end{cases}
\end{equation}
where $(u_n,v_n)$ is the coordinate of the $n$th point of the grid, $n=\{1,2,3,\ldots,N\}$ and $N$ is the total number of sampling, $\{ \cdot \}$ is the fractional part operator, e.g. $\{ 5.3 \} = 0.3 ,$ and 
$\mu$ is a grid parameter. 
Fig. \ref{miu_value} shows the effect of different $\mu$ values on the grid points. To make the grid points on the sphere as uniform as possible, the best value of $\mu$ is $\frac{\sqrt{5} -1}{2}(\approx 0.618) ,$ the reciprocal of the golden ratio.

After generating the Fibonacci grid, we project it onto 3D Cartesian coordinates on the unit sphere, obtaining uniformly distributed spherical grid points (Fig. \ref{fig_fibonacci}b):
\begin{equation}
\label{eq4}
    \begin{cases}
        \theta_n = \arccos(2 u_n - 1) \\
        \phi_n = 2\pi v_n
    \end{cases}
\end{equation}
where $\theta_n$ is the polar angle and $\phi_n$ is the azimuthal angle. 

We can now get the coordinate of grid points in the rectangular coordinate  (Fig. \ref{fig_fibonacci}c):
\begin{equation}
\label{eq4.1}
    \begin{cases}
        x_n = r\cos\phi_n\sin\theta_n \\
        y_n = r\sin\phi_n\sin\theta_n \\
        z_n = r\cos\theta_n
    \end{cases}
\end{equation}
where $x_n$, $y_n$, and $z_n$ are the coordinates of the point of the grid in the rectangular coordinate system, $r$ is a constant, chosen as $r=1$ to anchor a unit spherical surface.
\begin{figure}[!t]
\centerline{\includegraphics[width=\columnwidth]{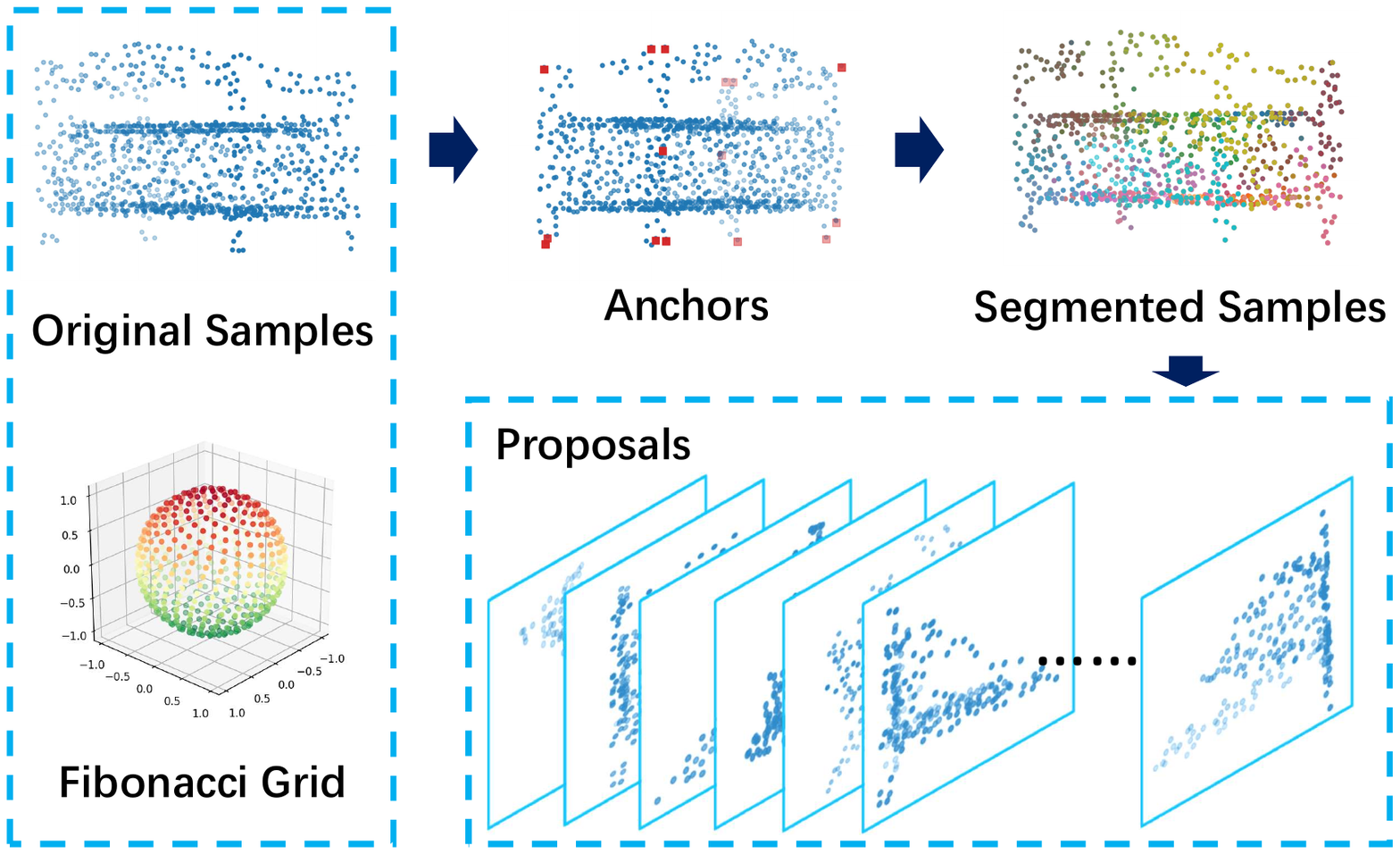}}
\caption{The process of segmenting the point clouds into proposals by the Fibonacci grid.}
\label{segmentation_process}
\end{figure}

After Eq. \eqref{eq4.1}, we have a set of uniform spherical anchor vectors $\vec{A} =\{ \vec{a}_n = (x_n,y_n,z_n) \mid n = 1, 2, 3,\ldots, N\} ,$ and can use these anchors to segment PCs by nearest search, as illustrated in Fig. \ref{segmentation_process}.
First, we translate the PC center to the origin by $\tilde{Q} = Q - \text{mean}(Q) .$ 
Then, we compute the dot product between the centered PC $\tilde{Q}$ and the anchor vector $\vec{a}_n$ and obtain the distance score by $c_n= \tilde{Q} \cdot \vec{a}_n .$ 
We get the indexes of the top-k highest scores by $I_n = \text{topk}(c_n)$ in Python and get the proposal $P_n=Q[I_n] .$ 
Finally, we obtain the segmented reference PC as $P = \{P_n \mid n = 1, 2,\ldots, N\} .$ 

After segmenting the PCs into patches, we can generate triplets from these proposals. Each triplet contains a source PC, a reference PC, and an irrelevant PC. We choose a proposal randomly as the source PC $S$ and the nearest proposal as the reference PC $Q .$ Next, we choose a proposal whose anchor is the farthest away from that of the source PC as the irrelevance PC $Ir ,$ completing the training triplet $(S, Q, Ir)$ selection.

\subsubsection{Model Training with Contrastive Learning}
With the established training triplet, we employ a shared feature extractor network, $f(\cdot)$, which processes each point cloud to generate its respective discriminative feature embedding: $F_S=f(S)$, $F_Q=f(Q)$, and $F_{Ir}=f(Ir)$. These learned features, along with the raw point clouds $S$ and $Q$, then feed into our pose estimation module, $g(\cdot)$, designed to directly regress the rigid transformation $\{R_{\text{pred}}, t_{\text{pred}}\}= g(S, Q, F_S, F_Q)$ to align $S$ with $Q .$

Our comprehensive training objective integrates two critical components to optimize both the accuracy of the pose estimation and the quality of the learned feature space. First, we minimize the error between our predicted transformation $\{R_{\text{pred}}, t_{\text{pred}}\}$ and the ground truth $\{R_{\text{gt}}, t_{\text{gt}}\}$ (e.g., via an $L_2$ or geodesic pose loss). Second, to sculpt a robust feature space, we enforce a triplet-based feature loss: minimizing the Euclidean distance $\parallel F_S - F_Q \parallel$ for positive pairs, while simultaneously maximizing the distance $\parallel F_S - F_{Ir} \parallel$ for negative pairs. This ensures our feature embeddings are inherently discriminative for accurate point cloud registration.

For the first goal, we leverage the registration loss of the baseline model:
\begin{equation}
\label{eq5}
    L_{\text{R}} = \text{loss}(R_{\text{gt}}, t_{\text{gt}},R_{\text{pred}},t_{\text{pred}})
\end{equation}
where $R_{\text{gt}}$,$t_{\text{gt}}$ are the GT transforms, and $R_{\text{pred}}$, $t_{\text{pred}}$ are the predicted transforms. The loss function quantifies the $SE(3)$ transformation error between the predictions and the ground truth, with its formulation adopted from the original literature of the baseline model. 
The precise formulation of the loss function is tailored to the baseline model's architecture and learning objectives. For example, in the RPMNet \cite{RPMNet} that we will use later in the experimentation section, the loss contains an $L_1$ distance and a weighted loss of the matching matrix:
\begin{equation}
\label{rpmloss}
\begin{aligned}
    L_{\text{RPM}} &= L_{\text{reg}} + \lambda_o L_{\text{inlier}} \\
    L_{\text{reg}} &= \frac{1}{J} \sum_{j}^{J} | ({R}_{\text{gt}} {s}_j + {t}_{\text{gt}}) - ({R}_{\text{pred}} {s}_j + {t}_{\text{pred}}) | \\
    L_{\text{inlier}} &=-\frac{1}{J} \sum_{j}^{J} \sum_{i}^{I} m_{ij} - \frac{1}{I} \sum_{i}^{I} \sum_{j}^{J} m_{ij}
\end{aligned}
\end{equation}

For our next goal of sculpting a robust feature space, we design a contrastive learning loss:
\begin{equation}
\label{eq6}
    L_{\text{Contrastive}} = \frac{1}{N} \sum_{n=1}^N \max(0, \parallel F_S - F_Q \parallel-\parallel F_S-F_{Ir} \parallel + \alpha)
\end{equation}
where $n$ is the size of the training set; $F_S, F_Q, F_{Ir} \in \mathbb{R}^d$ are the feature vectors of the source, the reference, and the irrelevance patches, respectively. The parameter $\alpha (> 0)$ is a margin that enforces a minimum separation between the feature embeddings of dissimilar samples, thereby preventing representational collapse where all embeddings converge to a single point.

The final loss function used for training the model is given by the following equation:
\begin{equation}
    L = L_{\text{R}} + \lambda L_{\text{Contrastive}}
\end{equation}
where $\lambda$ is a balancing coefficient to trade off the contributions of these two losses.

\begin{algorithm}[t]
\caption{Inference with Cascade Anchor Selection and Refinement (CASR)}
\label{alg:infer_casr}
\begin{algorithmic}[1]
\REQUIRE Source point cloud $S$; reference point cloud $Q$; baseline model $g(\cdot)$ and extractor $f(\cdot)$; number of anchors $N$; patch size $k$; EMA step $\eta$; iterations $K$.
\ENSURE Final rigid transform $(R^\star, t^\star)$.

\STATE Initialize Fibonacci anchors $\{\vec a_n^{0}\}_{n=1}^{N}$ on unit sphere.
\STATE $(R^\star, t^\star) \leftarrow (\mathbf{I}, \mathbf{0})$.
\FOR{$k = 0$ to $K-1$}
    \STATE \textbf{(Segment)} Center: $\tilde{Q} \leftarrow Q - \mathrm{mean}(Q)$.
    \FOR{$n = 1$ to $N$}
        \STATE Scores $c_n \leftarrow \tilde{Q}\cdot \vec a_n^{k} .$
        \STATE $I_n \leftarrow \mathrm{topk}(c_n)$, $P_n^{k} \leftarrow Q[I_n] .$
    \ENDFOR
    
    \STATE \textbf{(Register \& Evaluate)} 
    \FOR{$n = 1$ to $N$}
        \STATE $(R_n^{k}, t_n^{k}), \{\hat{q}_i\} \leftarrow g(S, P_n^{k}, f(S), f(P_n^{k}))$.
        \STATE Average distance $d_n^{k} \leftarrow \frac{1}{|S|}\sum_{i}\|R_n^{k}s_i + t_n^{k} - \hat{q}_i\|^2$ \hfill(Eq. \ref{eq7})
    \ENDFOR
    
    \STATE $b \leftarrow \arg\min_{n} d_n^{k}$; $(R^\star, t^\star) \leftarrow (R_b^{k}, t_b^{k})$.
    
    \FOR{$n = 1$ to $N$}
        \STATE $\vec a_n^{k+1} \leftarrow (1-\eta)\vec a_n^{k} + \eta \vec a_b^{k}$ \hfill(Eq. \ref{eq:update_anchor})
        \STATE Normalize $\vec a_n^{k+1} \leftarrow \vec a_n^{k+1}/\|\vec a_n^{k+1}\| .$
    \ENDFOR
\ENDFOR
\RETURN $(R^\star, t^\star)$.
\end{algorithmic}
\end{algorithm}

\subsection{Inference with Cascade Anchor Selection and Refinement}
In the inference stage, we segment the reference PC into proposals by the same method as in the training stage and utilize the trained baseline model to iteratively predict the transformations between the source PC and each individual proposal (Algorithm \ref{alg:infer_casr}).
In each iteration, we will select the best transform to update the anchors. Recall Eq. \eqref{eq1} and observe that a smaller value of cost indicates a better registration result. We employ the average distance as the evaluation metric. 
For each transform $\{R_n^k,t_n^k \} ,$ we calculate the average distance between the transformed source PC and the corresponding coordinates:
\begin{equation}
\label{eq7}
    d_n^k = \frac{1}{I} \sum_{i=1}^I \parallel R_n^k s_i+t_n^k-\hat{q}_i \parallel_2 
\end{equation}
where $s_i$ is a point in $S$ and $\hat{q}_i$ is the corresponding coordinate used in aligning $S$ and proposal $P_n$ to calculate each transform $\{R_n^k,t_n^k \}$. The transform with the lowest distance is then selected as the best result $\{R_b^k,t_b^k \} .$

After that, the anchor $\vec{a}_b^k$ related to the best result $\{R_b^k,t_b^k \}$ is used to update other anchors in the next iteration from an exponential moving average as shown in Eq. \ref{eq:update_anchor}:
\begin{equation}
\label{eq:update_anchor}
    \vec{a}_n^{k+1}=(1-\lambda)\vec{a}_n^k+ \gamma \vec{a}_b^k
\end{equation}
where $\vec{a}_n^k$ is the anchor of the current iteration, $\vec{a}_n^{k+1}$ is the anchor of the next iteration, and $\gamma$ is a hyperparameter chosen manually.

\section{Experimental Evaluation}
\label{sec:experimental evaluation}

\begin{table*}
\caption{Experimental results of R-SLPR and existing methods on ModelNet40 Dataset with the experimental setting "clean". The red fonts highlight the best result in each metric.}
\label{clean}
\centering
    \begin{adjustbox}{width=\textwidth}
    \begin{tabular}{c c c c c c c c c c }
    \hline
    \multirow{2}*{Experiment Set} & \multirow{2}*{Method} & \multicolumn{4}{c}{Sampling Rate 0.2} & \multicolumn{4}{c}{Sampling Rate 0.3} \\
    ~ & ~ & RMSE-r ($^\circ$) & MAE-r ($^\circ$) & RMSE-t & MAE-t & RMSE-r ($^\circ$) & MAE-r ($^\circ$) & RMSE-t & MAE-t   \\ \hline
           \multirow{9}*{Clean} & ICP \cite{besl1992method}   & 65.657 & 56.827 & 0.642 & 0.541 & 62.317 & 53.806 & 0.591 & 0.495  \\
           ~ & Small\_GICP \cite{Small_ICP}  &  64.329 & 54.128 & 0.587 & 0.501 & 61.334 & 52.152 & 0.547 & 0.422  \\
           ~ &  PointNetLK \cite{PointNetLK}  & 45.825 & 23.654 & 0.673 & 0.423 & 23.340 & 15.151 & 0.497 & 0.308 \\ 
           ~ &  PointNetLK-revisited \cite{PointNetLK-revisited}  & 26.916 & 17.365 & 0.567 & 0.349 & 25.230 & 16.238 & 0.544 & 0.294  \\ 
           ~ &  RPMNet \cite{RPMNet} & 4.102 & 2.223 & 0.065 & 0.020  & 3.841 & 2.011 & 0.039 & 0.026 \\ 
           ~ &  MFGNet \cite{MFGNet}  & 29.158 & 18.802 & 0.450 & 0.304 & 13.272 & 10.366 & 0.180 & 0.135 \\ 
           ~ &  RORNet \cite{RORNet}  & 18.635 & 15.314 & 0.322 & 0.240 & 16.469 & 12.391 & 0.340 & 0.271 \\ 
           ~ &  MCLNet \cite{MCLNet}  & 3.556 & 1.132 & 0.224 & 0.156 & 3.429 & 1.186 & 0.168 & 0.092 \\
           ~ &  \textbf{R-SLPR (ours)} & \textcolor{red}{2.632} & \textcolor{red}{1.104} & \textcolor{red}{0.020} & \textcolor{red}{0.009} & \textcolor{red}{3.304} & \textcolor{red}{1.268} & \textcolor{red}{0.027} & \textcolor{red}{0.012} \\
    \hline
\end{tabular}
\end{adjustbox}
\end{table*}

\begin{table*}
\caption{Experimental results of R-SLPR and existing methods on the ModelNet40 Dataset with the experimental setting "Unseen". The red fonts highlight the best result in each metric.
}
\label{Unseen}
\centering
\begin{adjustbox}{width=\textwidth}
    \begin{tabular}{c c c c c c c c c c }
    \hline
       \multirow{2}*{Experiment Set} & \multirow{2}*{Method} & \multicolumn{4}{c}{Sampling Rate 0.2} & \multicolumn{4}{c}{Sampling Rate 0.3} \\
       ~ & ~ & RMSE-r ($^\circ$) & MAE-r ($^\circ$) & RMSE-t & MAE-t & RMSE-r ($^\circ$) & MAE-r ($^\circ$) & RMSE-t & MAE-t   \\ \hline
      \multirow{9}*{Unseen} & ICP \cite{besl1992method}  & - & - & - & - & - & - & - & -  \\
       ~ & Small\_GICP \cite{Small_ICP} & - & - & - & - & - & - & - & - \\
       ~ &  PointNetLK \cite{PointNetLK}  & 44.203 & 27.126 & 0.591 & 0.365 & 21.200 & 16.312 & 0.400 & 0.269 \\ 
       ~ &  PointNetLK-revisited \cite{PointNetLK-revisited} & 27.320 & 16.602 & 0.452 & 0.300 & 25.540 & 16.231 & 0.411 & 0.278  \\
       ~ &  RPMNet \cite{RPMNet}  & 4.368 & 2.371 & 0.071 & 0.042 & 4.129 & 2.413 & 0.056 & 0.035 \\ 
       ~ &  MFGNet \cite{MFGNet}  & 30.556 & 20.222 & 0.237 & 0.173 & 11.521 & 7.072 & 0.090 & 0.060 \\ 
       ~ &  RORNet \cite{RORNet}  & 19.913 & 14.260 & 0.366 & 0.204 & 16.395 & 12.856 & 0.410 & 0.306  \\ 
       ~ &  MCLNet\cite{MCLNet}  & 3.638 & 1.328 & 0.150 & 0.098 & 3.521 & 1.200 & 0.140 & 0.101   \\
       ~ &  \textbf{R-SLPR (ours)} & \textcolor{red}{2.833} & \textcolor{red}{1.234} & \textcolor{red}{0.026} & \textcolor{red}{0.012} & \textcolor{red}{2.977} & \textcolor{red}{1.183} & \textcolor{red}{0.023} & \textcolor{red}{0.010} \\
         \hline
       
    \end{tabular}
\end{adjustbox}
\end{table*}

\begin{table*}
\caption{Experimental results of R-SLPR and existing methods on the ModelNet40 Dataset with the experimental setting "Noise". The red fonts highlight the best result in each metric. 
"-" means the value is not obtained for some reason.}
\label{Noise}
\centering
\begin{adjustbox}{width=\textwidth}
    \begin{tabular}{c c c c c c c c c c }
    \hline
       \multirow{2}*{Experiment Set} & \multirow{2}*{Method} & \multicolumn{4}{c}{Sampling Rate 0.2} & \multicolumn{4}{c}{Sampling Rate 0.3} \\
       ~ & ~ & RMSE-r ($^\circ$) & MAE-r ($^\circ$) & RMSE-t & MAE-t & RMSE-r ($^\circ$) & MAE-r ($^\circ$) & RMSE-t & MAE-t   \\ \hline
       \multirow{9}*{Noisy} & ICP \cite{besl1992method}   & 66.782 & 58.241 & 0.636 & 0.514 & 63.587 & 55.214 & 0.562 & 0.495  \\
       ~ & Small\_GICP \cite{Small_ICP}  &  65.327 & 56.831 & 0.638 & 0.499 & 64.521 & 55.368 & 0.582 & 0.433  \\
       ~ &  PointNetLK  \cite{PointNetLK} & 45.825 & 23.654 & 0.673 & 0.423 & 23.340 & 15.151 & 0.497 & 0.308 \\ 
       ~ &  PointNetLK-revisited  \cite{PointNetLK-revisited} & 26.394 & 17.575 & 0.507 & 0.388 & 25.894 & 14.238 & 0.519 & 0.352  \\ 
       ~ &  RPMNet  \cite{RPMNet} & 4.562 & 2.740 & 0.072 & 0.025  & 3.985 & 2.219 & 0.047 & 0.019 \\ 
       ~ &  MFGNet \cite{MFGNet}  & 33.340 & 21.842 & 0.404 & 0.320 & 14.395 & 9.567 & 0.168 & 0.124 \\ 
       ~ &  RORNet \cite{RORNet}  & 10.254 & 7.352 & 0.224 & 0.156 & 7.856 & 5.421 & 0.208 & 0.149  \\ 
       ~ &  MCLNet  \cite{MCLNet} & - & - & - & - & - & - & - & -   \\
       ~ &  \textbf{R-SLPR (ours)} & \textcolor{red}{2.613} & \textcolor{red}{1.142} & \textcolor{red}{0.027} & \textcolor{red}{0.013} & \textcolor{red}{2.974} & \textcolor{red}{1.229} & \textcolor{red}{0.021} & \textcolor{red}{0.010} \\
         \hline
       
    \end{tabular}
\end{adjustbox}
\end{table*}

\begin{table*}
\caption{Experimental results of R-SLPR and existing methods on the ModelNet40 Dataset with the experimental setting "Noise \& Unseen". The red fonts highlight the best result in each metric. "-" means the value is not obtained for some reason.
}
\label{Noise_and_Unseen}
\centering
\begin{adjustbox}{width=\textwidth}
\begin{tabular}{c c c c c c c c c c }
    \hline
       \multirow{2}*{Experiment Set} & \multirow{2}*{Method} & \multicolumn{4}{c}{Sampling Rate 0.2} & \multicolumn{4}{c}{Sampling Rate 0.3} \\
       ~ & ~ & RMSE-r ($^\circ$) & MAE-r ($^\circ$) & RMSE-t & MAE-t & RMSE-r ($^\circ$) & MAE-r ($^\circ$) & RMSE-t & MAE-t   \\ \hline
       \multirow{9}*{Noisy \& Unseen} & ICP \cite{besl1992method}   & - & - & - & - & - & - & - & -  \\
       ~ & Small\_GICP \cite{Small_ICP}  & - & - & - & - & - & - & - & - \\
       ~ &  PointNetLK \cite{PointNetLK}  & 46.014 & 25.236 & 0.594 & 0.317 & 23.721 & 14.106 & 0.443 & 0.357 \\ 
       ~ &  PointNetLK-revisited \cite{PointNetLK-revisited}  & 26.256 & 17.167 & 0.512 & 0.343 & 25.689 & 15.597 & 0.483 & 0.281  \\ 
       ~ &  RPMNet \cite{RPMNet}  & 4.714 & 2.828 & 0.041 & 0.022  & 4.114 & 2.263 & 0.033 & 0.019 \\ 
       ~ &  MFGNet \cite{MFGNet}  & 80.267& 61.713 & 0.424 & 0.348 & 14.952 & 10.697 & 0.127 & 0.084 \\ 
       ~ &  RORNet \cite{RORNet}  & 11.026 & 7.864 & 0.271 & 0.197 & 7.882 & 5.663 & 0.325 & 0.188  \\ 
       ~ &  MCLNet \cite{MCLNet}  & - & - & - & - & - & - & - & - \\
       ~ &  \textbf{R-SLPR (ours)} & \textcolor{red}{2.851} & \textcolor{red}{1.101} & \textcolor{red}{0.031} & \textcolor{red}{0.014} & \textcolor{red}{2.673} & \textcolor{red}{1.174} & \textcolor{red}{0.028} & \textcolor{red}{0.011} \\
         \hline
       
    \end{tabular}

\end{adjustbox}
\end{table*}

\begin{table}
\caption{\textcolor{black}{Ablation experiment results of R-SLPR on the ModelNet40 Dataset. The red fonts highlight the best result in each metric.}}
\label{Ablation_results}
\centering
\begin{adjustbox}{width=\columnwidth}
\begin{tabular}{ c c c c c c c }
\hline
    FG & CASR & CL & RMSE-r ($^\circ$) & MAE-r ($^\circ$) & RMSE-t & MAE-t   \\ \hline
    ~ & ~ & ~ & 4.102 & 2.223 & 0.065 & 0.020 \\ 
    \ding{51} & ~ & ~ & 3.268 & 1.016 & 0.035 & 0.012 \\ 
    \ding{51} & \ding{51} & ~ & 2.938 & 1.202 & 0.024 & 0.010  \\ 
    \ding{51} & \ding{51} & \ding{51} & \textcolor{red}{2.632} & \textcolor{red}{1.104} & \textcolor{red}{0.020} & \textcolor{red}{0.009} \\
       \hline

    \end{tabular}

\end{adjustbox}
\end{table}

\begin{table*}
\caption{Experimental results of R-SLPR and existing methods on the ISOPR Dataset with the experimental setting "Noise \& Unseen". The red fonts highlight the best result in each metric. "-" means the value is not obtained for some reason.
}
\label{ISOPR_results}
\centering
\begin{adjustbox}{width=\textwidth}
\begin{tabular}{c c c c c c c c c c }
    \hline
       \multirow{2}*{Experiment Set} & \multirow{2}*{Method} & \multicolumn{4}{c}{Sampling Rate 0.2} & \multicolumn{4}{c}{Sampling Rate 0.3} \\
       ~ & ~ & RMSE-r ($^\circ$) & MAE-r ($^\circ$) & RMSE-t & MAE-t & RMSE-r ($^\circ$) & MAE-r ($^\circ$) & RMSE-t & MAE-t   \\ \hline
       \multirow{5}*{Clean} & RPMNet \cite{RPMNet}  & 0.833 & 0.476 & 0.011 & 0.006 & 0.608 & 0.339 & 0.008 & 0.004 \\ 
       ~ &  MFGNet \cite{MFGNet}  & 15.118 & 10.424 & 0.208 & 0.143 & 11.032 & 7.134 & 0.141 & 0.095 \\ 
       ~ &  RORNet \cite{RORNet}  & 21.970 & 17.545 & 0.273 & 0.220 & 20.756 & 16.847 & 0.260 & 0.213  \\ 
       ~ &  MCLNet \cite{MCLNet} & 2.080 & 0.310 & 0.022 & 0.003 & 2.234 & 0.299 & 0.020 & 0.002\\
       ~ &  \textbf{R-SLPR (ours)} & \textcolor{red}{0.490} & \textcolor{red}{0.178} & \textcolor{red}{0.006} & \textcolor{red}{0.002} & \textcolor{red}{0.463} & \textcolor{red}{0.171} & \textcolor{red}{0.006} & \textcolor{red}{0.002} \\
       \hline
       \multirow{5}*{Noisy} & RPMNet \cite{RPMNet}  & 0.905 & 0.434 & 0.012 & 0.006 & 0.771 & 0.326 & 0.010 & 0.003 \\ 
       ~ &  MFGNet \cite{MFGNet} & 14.686 & 9.786 & 0.195 & 0.133 & 12.449 & 8.118 & 0.157 & 0.104 \\ 
       ~ &  RORNet \cite{RORNet} & 21.972 & 17.757 & 0.292 & 0.235 & 20.504 & 16.500 &  0.264 & 0.215  \\ 
       ~ &  MCLNet \cite{MCLNet} & - & - & - & - & - & - & - & - \\
       ~ &  \textbf{R-SLPR (ours)} & \textcolor{red}{0.506} & \textcolor{red}{0.225} & \textcolor{red}{0.006} & \textcolor{red}{0.002} & \textcolor{red}{0.498} & \textcolor{red}{0.211} & \textcolor{red}{0.006} & \textcolor{red}{0.003} \\
       \hline
    \end{tabular}

\end{adjustbox}
\end{table*}

\textcolor{black}{
To validate the  proposed method, we conduct comprehensive experiments on the general object dataset ModelNet40 \cite{ModelNet40} to evaluate its overall performance. 
We first introduce the ModelNet40 dataset and explain how it is processed for SLPR problem (Sec. \ref{sec51}). 
Then, we describe the experimental formulation, including the evaluation protocols, baseline methods, and quantitative metrics (Sec. \ref{sec52}). 
Finally, we present the experimental results and analyze the performance of the proposed method under different settings (Sec. \ref{subsec:ModelNet40_Results}).
Furthermore, we perform ablation studies on the key components of the proposed method to analyze their contributions to the overall performance and demonstrate their necessity (Sec. \ref{subsec:Ablation_Results}).
In addition, we verify its applicability to manufacturing scenarios on the industrial dataset ISOPR \cite{iLSPR} (Sec. \ref{subsec:ISOPR_Results}). 
}
\subsection{Dataset Introduction and Data Preparation}
\label{sec51}

ModelNet40 is a widely used benchmark in the fields of 3D computer vision and PC analysis. 
It consists of 12311 pre-aligned 3D CAD models. 
For each model, we extract a PC with 1024 points, whose coordinates are normalized into $[-1, 1]$. 
These CAD models are categorized into 40 common object classes, such as airplanes, chairs, and tables, which are formally divided into 9843 training samples and 2468 test samples, providing a standardized foundation for evaluating model generalizability. 

The ISOPR dataset is an industrial scene object point-cloud registration dataset constructed in NVIDIA Isaac Sim to evaluate point-cloud registration methods under manufacturing-oriented conditions. It contains 2,000 partial point clouds of 67 common workpiece models captured from simulated depth-camera observations, together with their ground-truth positions and orientations.

For the small-to-large registration, we extract a local point cloud from each sample as the source point cloud $S$, and use the original full sample as the reference point cloud $Q$. 
Then, we apply random transforms as ground truth transforms on the source point clouds and let the models predict the transforms.

\subsection{Experiment Formulation}
\label{sec52}
For the proposed R-SLPR, we choose RPMNet as the baseline model because of its excellent registration accuracy and robustness.
All implementations are developed in Python. Experiments are conducted on a workstation equipped with an NVIDIA RTX 4090 GPU and an Intel Core i9-14900K CPU (6.0 GHz) running Ubuntu 22.04. 
We employ the Adam optimizer with a learning rate of 0.001. 
The baseline model is trained by the method mentioned in Section 4.1.

To rigorously evaluate the proposed method, we conduct comprehensive comparisons against both handcrafted (e.g., ICP, Small-GICP) and representative learning-based (e.g., PointNetLK \cite{PointNetLK}, RPM-Net\cite{RPMNet}, MFGNet\cite{MFGNet}) point cloud registration approaches. 
All learning-based baselines are faithfully reproduced using their official implementations and the training protocols described in their respective publications. 
To evaluate the registration results, we compute the Root Mean Square Error (RMSE-r) and the Mean Absolute Error (MAE-r) of the Euler angles as metrics for rotation, and we compute the RMSE and MAE of the translation vector, labeled as RMSE-t and MAE-t, for translation \cite{iLSPR}. 

To demonstrate the generalizability of the proposed method, we conducted comprehensive experiments under varying configurations by following previous works: 1) clean, 2) unseen, 3) noise, and 4) noise and unseen. 
In the ``unseen'' setting, models are trained on the first 20 categories and tested on the last 20 categories, while, in the ``noise'' setting, Gaussian noise is added into both the training and test samples. 
Furthermore, to systematically evaluate the robustness of the proposed method to varying source PC densities, we conducted experiments at multiple cardinality ratios, specifically $\gamma = 0.2$ and $\gamma = 0.3 .$

\subsection{ModelNet40 Experiment Result}
\label{subsec:ModelNet40_Results}

\begin{figure}[!t]
\centerline{\includegraphics[width=\columnwidth]{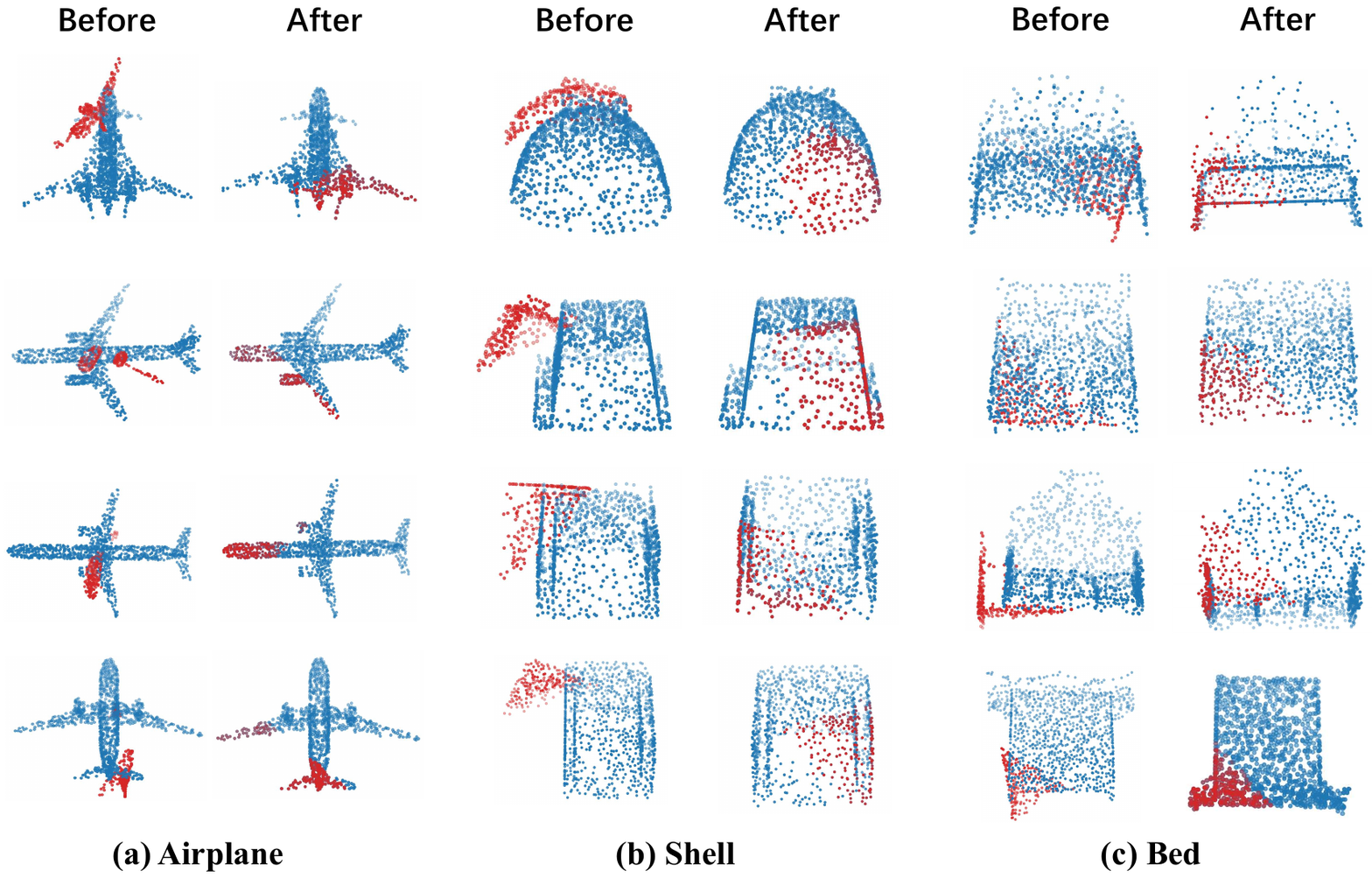}}
\caption{Example matching results of R-SLPR. The left columns display the point clouds before registration, while the right columns present the aligned point clouds.
}
\label{registration_result}
\end{figure}

Fig. \ref{registration_result} shows a few examples of the registration experiments.
Tables \ref{clean}, \ref{Unseen}, \ref{Noise}, and \ref{Noise_and_Unseen} summarize the experimental results under all configurations. Overall, R-SLPR consistently outperforms prior approaches across all conditions. 
In particular, it achieves robust performance with the RMSE-r below 3.0 degrees and the MAE-r below 1.3, while the RMSE-t is below 0.025. 
Among prior methods, RPMNet and MCLNet performed best, with  approximately RMSE-r 4.0 and 3.5, respectively. Other methods, such as MFGNet, PointNetLK, etc., exhibit poor fitting behavior when faced with the small-to-large registration problem. 
These results demonstrate that R-SLPR not only surpasses existing methods but also fulfills accuracy requirements for real-world applications. 

The comparative results demonstrate that the proposed method achieves superior performance in nearly all evaluation metrics. 
In the experimental settings ``Clean'' and ``Unseen'', MCLNet delivers the strongest performance among the baseline methods, attaining an RMSE-r of approximately 3.5, as shown in Tables \ref{clean} and \ref{Unseen}.
However, as shown in Tables \ref{Noise} and \ref{Noise_and_Unseen}, it exhibits notable limitations in noise robustness, failing to converge in both ``Noise'' and ``Noise \& Unseen'' configurations, which preclude reporting meaningful results for these cases. It is noteworthy that a noticeable discrepancy exists between the RMSE and MAE values of the proposed method. This phenomenon primarily stems from the higher sensitivity of RMSE to outlier predictions. In the context of the small-to-large registration problem, small PCs, randomly segmented from the large PCs, may be geometrically uninformative regions that lead to occasional registration failures, which inflate the value of RMSE compared to MAE. 

In terms of translation accuracy, the proposed method significantly outperforms all competing approaches, achieving RMSE-t from 0.02 to 0.03 among the experimental settings, while the RMSE-t of the best performing previous method, RPMNet, is from 0.4 to 0.7. This advantage stems from a fundamental limitation of existing approaches that they rely on the method mentioned in Section \ref{sec:slpcr} to compute translation by aligning PC centroids, which is inappropriate for the small-to-large registration where the centroids of the two point clouds are inherently misaligned. In contrast, R-SLPR first identifies corresponding regions in the large reference PCs and then estimates the rigid transforms, resulting in substantially higher precision in translation. 

When tested on unseen object categories, R-SLPR demonstrates robust overall performance and does not exhibit significant degradation in accuracy with (shown in Tables \ref{Unseen} and \ref{Noise_and_Unseen}) or without (shown in Tables \ref{clean} and \ref{Noise}) unseen objects. This resilience stems from its core design principle: rather than relying on features of PCs themselves, the method focuses on learning the relationship between the source PCs $S$ and the reference PCs $Q$. This fundamental emphasis on cross-instance relationships enables strong generalization capability, allowing the framework to maintain stable performance even when applied to object categories not encountered during training.

Similarly, R-SLPR demonstrates robust performance under Gaussian noise corruption. 
This resilience is attributed to the CASR method, which leverages the average distance between transformed source PCs and corresponding coordinates for proposal selection and can effectively mitigate noise interference. 
In contrast, several previous methods (e.g., MCLNet) exhibit significant performance degradation under noisy conditions, with some even failing to converge properly due to their sensitivity to perturbed point distributions.

\subsection{Ablation Study}
\label{subsec:Ablation_Results}

To quantify the effectiveness of the major components in R-SLPR, we conduct a comprehensive ablation study under the same small-to-large point-cloud registration protocol used in the main experiments. 
Because the proposed framework integrates region proposal generation, contrastive regional representation learning, and cascade anchor selection and refinement, some components are structurally necessary for maintaining a valid registration pipeline and cannot be simply removed. 
Accordingly, we evaluate the contribution of these designs through both component removal and controlled replacement. 
In particular, we compare the full R-SLPR framework with variants that remove the contrastive learning loss, disable the CASR refinement module, or replace the Fibonacci Grid Segmentation strategy with alternative anchor-generation schemes such as random sampling or farthest point sampling. In addition, we investigate the influence of the number of CASR iterations to assess the role of iterative proposal refinement. 
All ablated variants are evaluated using the same training and testing splits, source-to-reference ratios, and performance metrics as the main benchmark experiments. 
This ablation study aims to determine whether the observed performance gains arise from the proposed region-based formulation and its individual algorithmic components rather than from the registration backbone alone.

\subsection{ISOPR Experiment Result}
\label{subsec:ISOPR_Results}

To further demonstrate the applicability of R-SLPR in manufacturing scenarios, we evaluate it additionally on the ISOPR dataset proposed in the literature\cite{iLSPR}. 
Different from ModelNet40, ISOPR contains point clouds of industrial parts with more manufacturing-specific geometric characteristics, such as local planar surfaces, sharp edges, repetitive structures, and partial geometric ambiguity. 

Since the ISOPR experiment is intended as an application-oriented validation rather than a full-scale benchmark, we compare R-SLPR with representative baselines selected from the ModelNet40 experiments, including MFGNet~\cite{MFGNet}, RPMNet~\cite{RPMNet}, RORNet~\cite{RORNet}, and MCLNet~\cite{MCLNet}. 
The experimental formulation is consistent with the protocols used in previous studies~\cite{iLSPR} and follows the configurations described in Sec.~\ref{sec52}.

The results are reported in Table \ref{ISOPR_results}. 
Overall, R-SLPR achieves the best performance on the ISOPR dataset, indicating that the proposed region-based formulation generalizes well to manufacturing-style point clouds. 
Compared with RPMNet, R-SLPR reduces the rotation RMSE from 0.833$^\circ$ to 0.490$^\circ$ and the translation RMSE from 0.011 to 0.006 under the clean setting with a sampling rate of 0.2. 
Compared with MCLNet, R-SLPR further improves the rotation MAE from 0.310$^\circ$ to 0.178$^\circ$ and the translation MAE from 0.003 to 0.002. 
These improvements suggest that directly applying conventional registration methods to small-to-large manufacturing data remains challenging, especially when the source point cloud covers only a local region of the reference model.

The superior performance of R-SLPR mainly comes from its explicit region identification mechanism. 
By first localizing the most plausible corresponding region and then performing refined registration with CASR, R-SLPR effectively reduces the influence of scale mismatch and centroid inconsistency in small-to-large registration. 
The ISOPR results further show that the proposed method can generalize beyond synthetic ModelNet40 benchmarks and is applicable to manufacturing-oriented tasks such as CAD-to-scan alignment, workpiece localization, and robotic inspection.

\section{Conclusion and Future Work}
\label{sec: conclusion and future work}



This paper presented Region-based Small-to-Large Point-cloud Registration (R-SLPR), a framework designed to overcome significant scale discrepancies in point-cloud alignment through a multi-stage regional approach. By integrating Fibonacci Grid Segmentation (FGS) with a contrastive learning-based model and a Cascade Anchor Selection and Refinement (CASR) algorithm, the method effectively bridges the gap between small-scale source patches and large-scale reference environments. Experimental validation on the ModelNet40 dataset confirms the precision of this approach, yielding a position Mean Absolute Error (MAE) of 0.009 and a rotation MAE of 1.104°. These results demonstrate that the proposed region-based formulation significantly enhances registration robustness and estimation accuracy in ``small-to-large'' settings.

This research opens up several promising directions for future work. First, an end-to-end framework could be developed to unify proposal generation, selection, and registration within a single learnable pipeline, potentially enhancing both efficiency and coordination between modules. Also, the proposal screening and generation mechanism could benefit from further optimization such as the introduction of a hierarchical scoring network or more geometrically-aware sampling strategies. 

\end{document}